\documentclass[]{amap}

\usepackage{booktabs}
\usepackage{multirow}
\usepackage{graphicx}
\usepackage{array}
\usepackage{siunitx}
\usepackage[table]{xcolor}
\usepackage{makecell}
\usepackage{framed}
\usepackage{hyperref}
\usepackage{amsmath}
\usepackage{amsfonts}
\usepackage{placeins}
\usepackage{longtable}
\usepackage{hhline}
\usepackage{fancyvrb}
\usepackage{float}
\usepackage{fvextra}
\usepackage{multicol}
\usepackage{cleveref}
\usepackage{tablefootnote}
\usepackage{threeparttable}
\usepackage{tabularx}
\usepackage{mdframed}
\usepackage[usestackEOL]{stackengine}
\usepackage[numbers]{natbib}
\usepackage{enumitem}
\usepackage{adjustbox}
\usepackage{arydshln}
\usepackage{pifont}
\usepackage[dvipsnames]{xcolor}

\newcommand{\commentout}[1]{}
\renewcommand{\paragraph}[1]{\noindent\textbf{#1.}\hspace*{1em}}
\setlist[itemize]{leftmargin=15pt}

\title{PhysBrain 1.0 Technical Report}

\author{PhysBrain Team}

\contribution{See \hyperref[sec:contributions]{Contributions section} for a full author list.}

\projectpage{https://phys-brain.github.io/}

\abstract{Vision-language-action models have advanced rapidly, but robot trajectories alone provide limited coverage for learning broad physical understanding. PhysBrain 1.0 studies a complementary route: converting large-scale human egocentric video into structured physical commonsense supervision before robot adaptation. Our data engine extracts scene elements, spatial dynamics, action execution, and depth-aware relations, then turns them into question-answer supervision for training PhysBrain VLMs. The resulting physical priors are further transferred to VLA policies through a capability-preserving and language-sensitive adaptation design. Across multimodal QA benchmarks and embodied control benchmarks, including ERQA, PhysBench, SimplerEnv-WidowX, LIBERO, and RoboCasa, PhysBrain 1.0 achieves SOTA results and shows especially strong out-of-domain performance on SimplerEnv. These results suggest that scaling physical commonsense from human interaction video can provide an effective bridge from multimodal understanding to robot action.
}

\begin{document}
\newcommand{\todo}[1]{\textcolor{red}{TODO: #1}}
\newcommand{\lz}[1]{\textcolor{blue}{Lyna: #1}}
\newcommand{\infobox}[1]{
    \begin{tcolorbox}[
        width=0.97\linewidth,
        center,
        colback=amappaleblue!45!white,
        colframe=amapblue!75!black,
        arc=5pt,
        boxsep=4pt,
        left=4pt,
        right=4pt,
        top=3pt,
        bottom=4pt,
        boxrule=0.8pt,
    ]
        \begin{minipage}{0.98\linewidth}
        \small\raggedright
        #1
        \end{minipage}
    \end{tcolorbox}
}

\maketitle

\begin{figure}[H]
	\centering
    \vspace{-3em}
    \includegraphics[width=\textwidth]{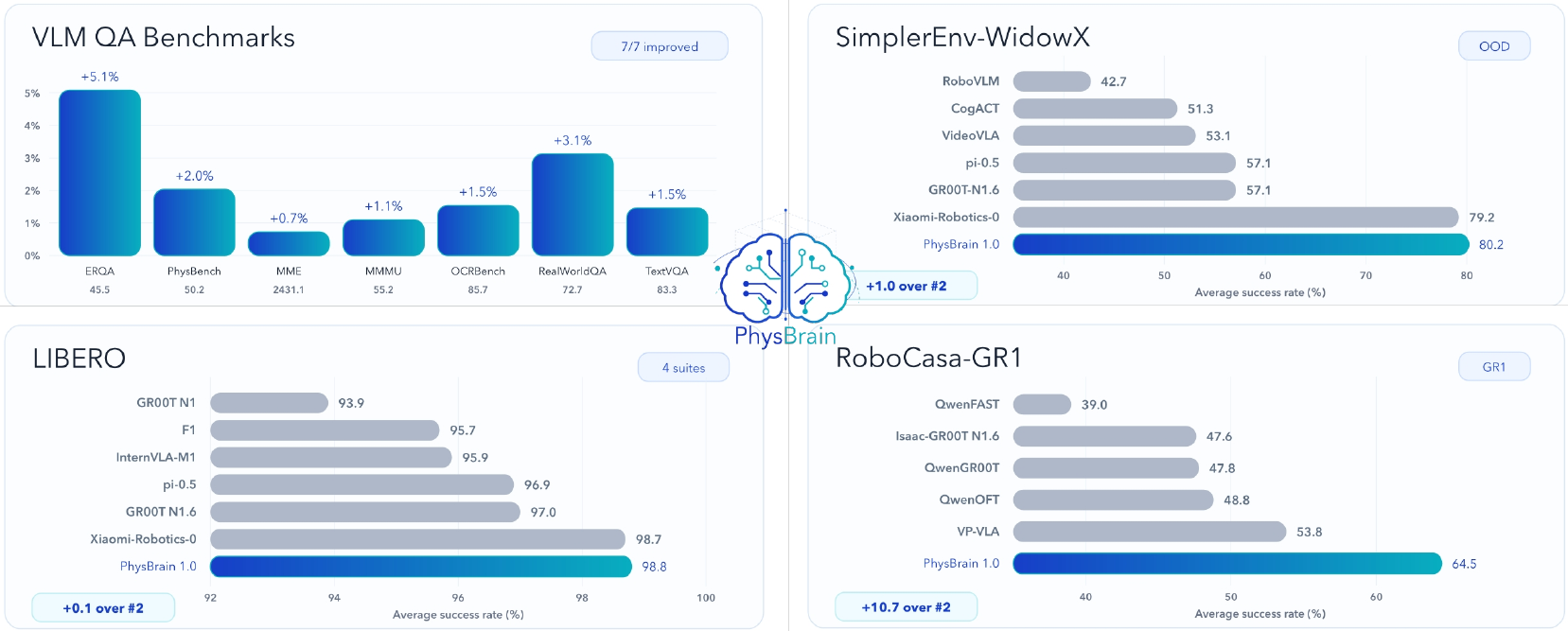}
	\caption{\textbf{PhysBrain 1.0 overall system overview.} PhysBrain 1.0 transforms large-scale human egocentric interaction videos into structured physical supervision, including scene elements, spatial dynamics, action execution, and depth- aware relations, and renders these records into physically grounded QA for training a stronger base VLM. The learned physical priors are then transferred to robot control through capability-preserving VLA adaptation, supporting language-conditioned action generation across simulated and real-world embodied tasks.}
\end{figure}

\clearpage
\tableofcontents
\clearpage

\section{Introduction}
\label{sec:intro}

\vspace{1mm}
\begin{quote}
\raggedright
\textit{``Understanding first, action next.''}\par
\vspace{-0.4em}
\raggedleft
\textit{--- Core principle of PhysBrain 1.0}
\end{quote}

Recent vision-language-action (VLA) systems have shown that large multimodal models can be adapted to robot control, but much of the field is still organized around one dominant training logic: collect robot trajectories, fit action policies, and scale the system by increasing the amount of robot interaction data. This route has produced important progress, yet it also narrows the source of embodied capability to expensive, platform-dependent trajectory collection. More importantly, fitting trajectories alone does not guarantee that the model has learned the physical regularities that support robust action under changes in viewpoint, scene layout, object state, or task composition.

PhysBrain 1.0 explores a different premise. We argue that embodied intelligence training should move from \textbf{action imitation} toward \textbf{physical commonsense acquisition}. Rather than scaling a more general embodied policy purely through robot trajectories, PhysBrain 1.0 first builds a general multimodal model with stronger physical understanding, and only then adapts it to embodied control.

This shift in training logic also requires a different source of data. To move beyond expensive human-teleoperated robot trajectories whose coverage is limited by platform, scene diversity, and collection budget, PhysBrain 1.0 turns to large-scale human first-person video as an alternative source of supervision. Compared with robot datasets, egocentric human video is easier to obtain, broader in coverage, and naturally centered on interaction with the physical world. It repeatedly exposes contact, reachability, object state change, tool use, spatial constraint, and multi-step task structure. These patterns are closely aligned with the kinds of physical regularities that VLA systems must ultimately reason about. \textbf{This report therefore focuses on two connected questions: whether human first-person video can be systematically transformed into scalable physical supervision, and whether the resulting priors can transfer effectively to downstream embodied control.}

Human first-person data are promising, but raw human video is not yet embodied supervision. By itself, it does not provide the explicit signals that a model can directly use for physical reasoning and action-oriented understanding. To address the first question, PhysBrain 1.0 introduces a schema-driven data annotation pipeline that first extracts structured scene meta-information and then uses it to generate physically grounded QA. The central design choice is to make the latent physical factors explicit before supervision is produced: what objects are present, how they are arranged, how their spatial relations evolve during manipulation, which actions are physically feasible, and how local execution supports a broader task objective. In this sense, the data engine compiles video into meta records over scene elements, spatial dynamics, execution process, and depth-aware relations, and then turns those records into natural-language question-answer supervision.

Once this data engine has been used to construct large-scale supervision and train a stronger base VLM, the second question becomes how to transfer those physics-based priors effectively into downstream robot control. Prior VLM-to-VLA studies have already shown both the opportunity and the risk of this route: multimodal models can be adapted into robot policies, but imitation-dominated post-training can also erode the original vision-language capability and lead to catastrophic forgetting~\cite{ChatVLA2_2025_arXiv,VLM2VLA_2025_arXiv,TwinBrainVLA_2026_arXiv}. PhysBrain 1.0 addresses this problem by assigning robot trajectories a narrower and more deliberate role. They remain important, but they are not treated as the sole source of embodied capability. Instead, the model first acquires stronger physical understanding from human interaction data, and then uses a limited amount of robot data for embodiment-specific adaptation. The architecture is designed accordingly: it preserves a stable general pathway during VLA training, keeps control sensitive to language rather than collapsing into a purely visual shortcut, and layers robot adaptation on top of a model that already carries stronger physical priors.

Empirically, this training logic yields strong results on both multimodal understanding and embodied control benchmarks. PhysBrain 1.0 performs well on ERQA~\cite{GoogleRobotics_2025_arxiv}, PhysBench~\cite{PhysBench_2025_arXiv}, MME~\cite{MME_2023_arXiv}, MMMU~\cite{MMMU_2024_CVPR}, OCRBench~\cite{OCRBenchV2_2025_arXiv}, RealWorldQA~\cite{realworldqa2024}, and TextVQA~\cite{TextVQA_2019_CVPR} on the VLM side, and on SimplerEnv-WidowX, SimplerEnv-GoogleRobot~\cite{SimplerEnv_2024_CoRL}, LIBERO~\cite{LIBERO_2023_NeurIPS}, and RoboCasa-GR1~\cite{RoboCasa_2024_RSS, GR00T_2025_arXiv} on the VLA side. Our main contributions are fourfold. First, we present a scalable annotation pipeline that transforms human first-person interaction video into structured scene meta-information and physically grounded QA rather than generic free-form captions. Second, we show that this supervision improves first-person embodied understanding in the base VLM by explicitly training perception, state, planning, and execution reasoning. Third, we introduce an integrated adaptation architecture that transfers these priors into downstream robot control while preserving useful general multimodal capability and language alignment. Fourth, we demonstrate that stronger human-derived priors can support strong downstream embodied performance using only limited benchmark-specific robot adaptation data.

\section{PhysBrain 1.0 Data Engine}
\label{sec:data_engine}

\subsection{Design Goal}

The PhysBrain 1.0 data engine is designed to answer a specific question: how can human first-person interaction video be converted into supervision that is useful for robot-oriented physical understanding? A naive answer would be to attach captions to video clips and ask the model to imitate those descriptions. We do not follow that route. Generic captions are too weak for embodied learning because they tend to summarize appearance or high-level events while leaving out the physical structure needed for action generation, such as object geometry, contact progression, relative distance, reachability, or the order of sub-actions.

Accordingly, the data engine is built around two principles. First, the supervision must be \textbf{physically explicit}. PhysBrain 1.0 makes this explicitness operational by first extracting structured \textbf{scene meta-information} from video: the records describe not only what is visible, but also which objects are present, what physical attributes they have, how they are spatially arranged, how depth relations are formed, and how the scene changes under action. Second, the pipeline must separate this \textbf{scene meta-information} from \textbf{model supervision}. The intermediate annotations are structured because they serve as source records for downstream generation in a machine-readable form. The final VLM training data, however, are still natural question-answer pairs. This separation lets PhysBrain 1.0 control the physical content of the data without reducing the model's training target to rigid JSON fields.

This design makes the data engine closer to a compiler than to a caption generator. Raw video is first parsed into an explicit physical record; the record is then augmented, checked, and finally rendered into QA supervision. Each stage has a constrained input-output interface, so errors can be detected before they propagate into the final training set.

\subsection{Data Sources and Staged Construction}

The training corpus for PhysBrain 1.0 is assembled in stages rather than from a single static dataset. The first stage focuses on egocentric sources such as Ego4D~\cite{Ego4D_2022_CVPR}, BuildAI~\cite{buildaiegocentric10k2025}, and EgoDex~\cite{EgoDex_2025_arXiv}, where clips are segmented from first-person human interaction videos and converted into structured scene meta-information. Before annotation, clips are filtered with both visual-quality scores and camera-motion scores. In practice, camera motion is estimated from VGGT-derived camera parameters~\cite{VGGT_2025_CVPR} and summarized as a motion score; segments with sufficient visual quality and bounded camera shake are retained, while low-quality or unstable clips are removed before meta-information extraction. The second stage expands the re-annotation process to sources such as EPIC~\cite{damen2020epic}, and SEA-Small~\cite{spatial_ai_sea_small}, with a stronger emphasis on physical reasoning: the objective is no longer only to identify what action occurs, but to organize the clip into objects, physical properties, spatial relations, depth cues, state changes, and action-relevant dynamics. A later stage uses these meta-information records to generate free-form VQA supervision across capability families, including depth-aware spatial reasoning, temporal understanding, embodied planning, fine-grained perception, and general multimodal reasoning. In addition, general multimodal data such as FineVision are mixed during training as auxiliary retention data rather than re-labeled from scratch.

This staged construction matters for the final narrative. PhysBrain 1.0 does not treat all human data as interchangeable. Different subsets serve different roles: scene meta-information extraction makes the physical content explicit, depth augmentation enriches 3D and metric spatial grounding, QA generation turns the extracted source information into trainable natural-language supervision, and general-purpose multimodal data help preserve broad vision-language competence. Together they form a curriculum for physical commonsense injection rather than a flat collection of video descriptions.

\subsection{Structured Scene Meta-Information}

The first layer of annotation is not used as direct VLM supervision. Instead, PhysBrain 1.0 first extracts structured scene meta-information from each video segment. Each segment is represented by a small set of uniformly sampled frames and processed with a constrained prompt that asks for JSON output only. The output schema has three top-level fields: \texttt{scene\_elements}, \texttt{spatial\_dynamics}, and \texttt{action\_execution}. These fields form the source record from which later QA examples are generated, and their structured format also makes automatic parsing and validation possible. To improve both quality and diversity, scene meta-information is annotated and cross-checked with a strong multi-model pool, including GPT-5, Gemini 3.1 Pro, Gemini 3 Pro, Qwen3-VL-235B-A22B, and Qwen3.5-397B-A17B. Using multiple annotators reduces the risk that the physical supervision collapses into the style, omissions, or reasoning biases of a single model, and helps expose the base VLM to a broader distribution of physically grounded descriptions.


\paragraph{Scene elements}
The \texttt{scene\_elements} field captures the static or slowly varying aspects of the clip that are most relevant to interaction. It identifies the main manipulated object, other nearby objects, visual details, and the surrounding environment. Importantly, these visual details are not generic appearance tags. The schema explicitly records material cues, geometry, and physical state, such as whether an object appears folded, scattered, transparent, rigid, or filled. This choice reflects the observation that physical feasibility often depends on such attributes. A graspable rigid handle, a deformable cloth, and a pile of loose small parts require different embodied interpretations even if they occupy similar image regions.

\paragraph{Spatial dynamics}
The \texttt{spatial\_dynamics} field records how the scene is laid out at the beginning of the clip and how the relation between actor and objects changes over time. The annotation prompt asks for an \texttt{initial\_layout} and a \texttt{spatial\_change} description. This turns the supervision from static recognition into physically situated change modeling. Instead of merely saying that a hand interacts with an object, the annotation specifies whether the hand approaches from above, closes distance until contact, separates a part from a pile, reorients an object, or shifts it relative to a support surface.

\paragraph{Action execution}
The \texttt{action\_execution} field contains two complementary views of the task: a short \texttt{instruction\_brief} and a more detailed \texttt{execution\_detailed}. The brief instruction serves as the compact task intent. The detailed execution expands it into an imperative sequence emphasizing trajectory, velocity profile, and contact physics. This makes the output more useful than plain narration because it explicitly links the observed motion to an actionable control description.

Taken together, these three fields move the annotation process beyond simple captioning. They separate object identity from spatial relation and execution process, which gives the next stage a reliable physical basis for generating diverse QA. 

\subsection{Depth-Aware Spatial Augmentation}

Structured scene meta-information alone is still limited when the task requires 3D relation or depth-sensitive planning. To address this, PhysBrain 1.0 adds a depth-aware spatial augmentation stage. For clips with object grounding metadata, the pipeline associates scene objects with point-wise depth estimates computed by Depth Anything v3~\cite{lin2025depth}, using the DA3NESTED-GIANT-LARGE-1.1 depth model. In practice, the pipeline locates each object's center point, rescales it into the depth-map coordinate system, and records a compact \texttt{depth\_info} dictionary for the clip.

This augmentation serves two purposes. First, it supports \textbf{relative depth} QA, where the model learns whether an object is closer, farther, behind, lower, or more reachable than another object. Such questions help the VLM distinguish semantic co-occurrence from physical arrangement. Second, it supports \textbf{absolute depth} and metric-distance QA, where the model learns real-world distance and scale in meters or centimeters. This matters for downstream action generation because some robot demonstration data are represented through end-effector positions, poses, or displacements. A model that has learned only ordinal relations may know which object is nearer, but a model exposed to metric depth supervision has a better basis for understanding absolute position and continuous spatial displacement.

Depth-aware augmentation therefore gives the data engine a concrete way to encode both ordinal 3D layout and metric spatial structure. The final answers remain natural language QA, but their generation is grounded in explicit depth metadata rather than visual appearance alone. Invalid or missing depth records can be identified at this intermediate stage, before they are used to construct spatial QA.

\subsection{QA Generation}

The third layer is QA generation. This is the stage that turns structured scene meta-information into the actual VLM training examples. The role of the upstream metadata is to make the generated QA physically grounded: questions can ask about objects, physical properties, spatial relations, depth, state changes, feasible actions, and long-horizon plans because those factors have already been extracted from the source video. QA generation uses the full multi-model pool, including GPT-5, GPT-5 mini, Gemini 3.1 Pro, Gemini 3 Pro, Qwen3-VL-30B-A3B, Qwen3-VL-235B-A22B, Qwen3.5-35B-A3B, and Qwen3.5-397B-A17B. Different annotator models tend to phrase questions differently, emphasize different physical cues, and expose different reasoning paths. This helps prevent the trained VLM from inheriting the narrow supervision style of any single generator and mitigates a potential performance bottleneck caused by homogeneous synthetic labels.

\begin{figure*}[t]
    \centering
    \includegraphics[width=\textwidth]{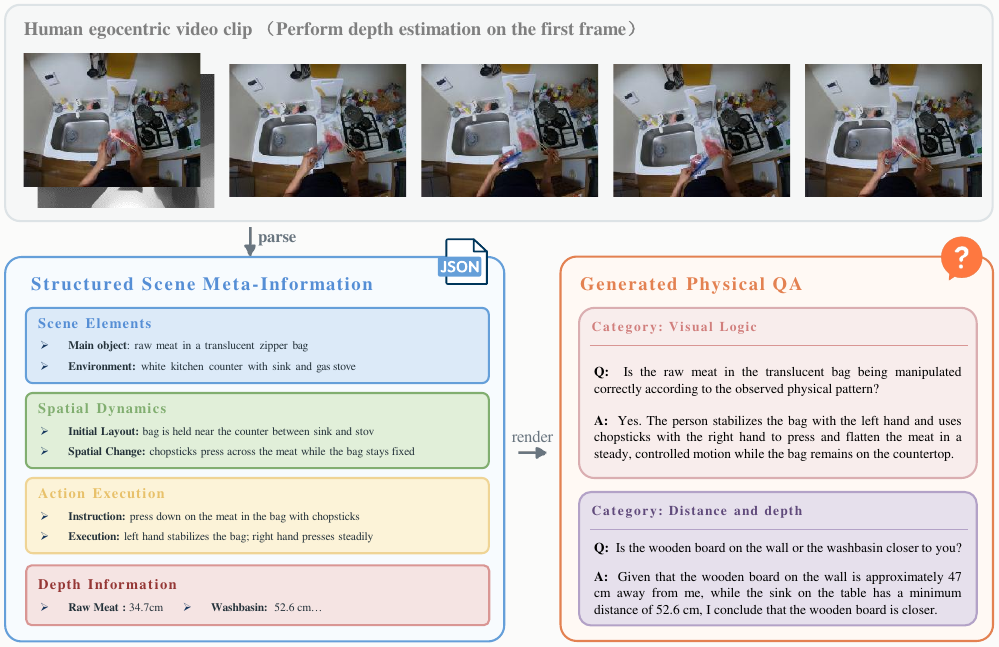}
    \caption{\textbf{Example of structured meta-information and generated physical QA.}
    We uniformly sample from an egocentric manipulation clip and convert the clip into a compact JSON-style source record. The record separates static scene elements, spatial changes, and action execution details, which are then used to generate physically grounded QA supervision.}
    \label{fig:meta_qa_example}
\end{figure*}

Figure~\ref{fig:meta_qa_example} shows a representative instance of this conversion process. A short egocentric clip is first represented by uniformly sampled frames, then parsed into structured meta-information over scene elements, spatial dynamics, and action execution. The final QA example is rendered from this source record.

{\footnotesize
\setlength{\tabcolsep}{3pt}
\renewcommand{\arraystretch}{1.3}
\begin{longtable}{@{}p{0.27\linewidth}p{0.49\linewidth}p{0.21\linewidth}@{}}
\caption{Capability coverage of the PhysBrain 1.0 QA generation stage. The table summarizes the QA families used to
convert structured scene meta-information into trainable natural-language supervision.}\\
\toprule
\textbf{QA family} & \textbf{Main target} & \textbf{Training role} \\
\midrule
Spatial relations & Left/right, above/below, and front/behind relations & Spatial intelligence \\
Distance and depth & Relative depth and absolute metric distance & Spatial grounding \\
Size estimation & Real-world length, width, height, and object scale & Metric understanding \\
Grounding and coordinates & Bounding boxes, points, and vacant-space coordinates & Visual grounding \\
Viewpoint reasoning & Cross-view consistency and object-facing direction & Egocentric reasoning \\
Next-step prediction & Action choice under the current observation and goal & Embodied decision making \\
Route planning & Navigation direction and route completion & Embodied navigation \\
Affordance and safety & Operability, touch safety, and immediate danger & Physical commonsense \\
Long-horizon planning & Multi-step task decomposition & Long-horizon control \\
Object state change & Physical outcome after manipulation & Dynamics modeling \\
Action recognition and counting & Performed action and repetition count & Video understanding \\
Temporal ordering & Event order and object appearance order & Temporal reasoning \\
Action localization & Time interval of a specified action & Video grounding \\
Causal/counterfactual reasoning & Why-events and what-if outcomes & Physical causality \\
Counting & Object counts and attribute-conditioned counts & Fine-grained perception \\
Fine-grained attributes & Material, color, state, height, and reflectance & Attribute recognition \\
Existence checking & Whether an object appears, or appears only at certain times & Hallucination suppression \\
Scene text and OCR & Signs, labels, screens, prices, and dates & General retention \\
Chart and data analysis & Charts, arithmetic, and geometric quantities & General retention \\
Science and technical knowledge & Physics, chemistry, circuits, and domain problems & Knowledge retention \\
Visual logic & Pattern completion, Raven-style reasoning, and forensics & Abstract reasoning \\
\bottomrule
\label{tab:qa_families}
\end{longtable}
}

The QA space is organized around capability families rather than around a single benchmark. Some categories preserve general multimodal capability, including document, chart, OCR, counting, professional knowledge, and visual logic questions. Some categories target spatial intelligence, including 3D spatial relations, metric distance, size estimation, visual grounding, viewpoint reasoning, and geometric reasoning in the spirit of recent work on surrogate geometric tasks~\cite{Euclids_Gift_2025_arXiv}. Other categories target embodied capability, including next-step prediction, route planning, affordance and safety, long-horizon task decomposition, and object state change. Temporal categories further train the model to understand action order, action localization, and causal or counterfactual dynamics.

QA generation is not intended to produce a single natural-language description of each video. Instead, it uses the same physical scene meta-information to instantiate multiple forms of supervision, so that a single interaction clip can support questions about spatial layout, metric depth, object state, future actions, safety, temporal order, and high-level reasoning. This is one of the clearest differences between PhysBrain 1.0 and unconstrained video caption pipelines: the final QA can remain linguistically diverse and free-form, while the physical content is controlled by the structured metadata that precedes it.

\subsection{From Annotation Schema to Embodied Reasoning Targets}

To strengthen the PhysBrain base model's first-person physical reasoning ability, QA answers follow a principled embodied reasoning format when the task involves physical interaction, planning, or action feasibility:

\infobox{
\begin{quote}
\small
[Perception - Environment] $\rightarrow$ [Perception - Object] $\rightarrow$ [Spatial Planning] $\rightarrow$ [Action Execution].
\end{quote}
}
In this format, the model is first asked to identify the environment, then characterize the manipulated object and its physical state, then reason about the spatial layout and its intended change, and only after that describe the concrete execution. The corresponding prompt explicitly frames the model as an embodied agent and asks it to analyze the environment, objects, and spatial dynamics step by step before detailing motor execution.

This training target is central to the PhysBrain 1.0 formulation. The goal is not to teach the model to produce longer answers, but to force an internal ordering of thought that is aligned with embodied action: perceive, infer state, plan, execute. In other words, PhysBrain supervision is structured to encourage \emph{physical organization} of the scene before action generation. This differs from generic instruction tuning, where a model may answer correctly while bypassing the intermediate physical factors that matter for control transfer.

\subsection{Quality Control and Noise Suppression}

Quality control is applied at the interfaces between annotation stages rather than only as a final cleanup step. Each stage produces an intermediate record with required fields, parseable structure, and explicit status information. This design does not make large-scale video annotation noise-free, but it turns many common failures into detectable cases: missing frames, invalid JSON, incomplete records, absent depth files, unreadable depth maps, or examples without valid object grounding.

Before scene meta-information extraction, clips are filtered by segment quality and motion scores so that low-information or visually unstable segments are less likely to enter the annotation pool. During extraction, the annotator is constrained to fill a fixed set of evidence-oriented fields, including \texttt{scene\_elements}, \texttt{spatial\_dynamics}, and \texttt{action\_execution}. The role of these constraints is not to assume that a textual instruction such as ``do not hallucinate'' is sufficient. Rather, the schema narrows what can be accepted as a valid intermediate record: the output must be parseable as JSON, must contain the expected fields, and must express physical content through visible scene elements, spatial changes, and action execution. Records that fail parsing, lack usable images, exceed generation limits, or return extraction errors are assigned failure statuses instead of being silently passed to QA generation.

Depth processing adds another set of checks at the object-grounding interface. For each grounded object, the pipeline verifies that the corresponding depth file exists, that the sampled image exists, and that the depth array can be loaded. It then reads the original image size and the depth-map shape, maps the object center from image coordinates into depth-map coordinates using the image-to-depth scale factors, bounds the resulting index inside the depth map, and samples the depth value at the mapped location. If the depth file is missing, the source image is missing, or the depth array cannot be loaded, the example is written with sentinel depth values and a non-success \texttt{depth\_status} such as \texttt{npz\_missing}, \texttt{image\_missing}, or \texttt{npz\_corrupted}. Downstream QA generation can then avoid depth-dependent questions for these examples while still using other valid scene information when appropriate.

These checks reduce the probability that malformed or unsupported intermediate artifacts become final supervision. They do not eliminate all semantic noise or all depth-estimation errors, but they make the failure modes more visible and easier to filter. As a result, the final QA can remain free-form and linguistically diverse, while its physical content is tied to source records that have passed a set of structural and modality-specific checks.

\section{PhysBrain 1.0 Architecture}
\label{sec:architecture}

\subsection{Overview}

After building a general multimodal model with stronger physical understanding, the next step is to adapt this model to action. To support this transition, PhysBrain 1.0 inherits two design lines from prior work: a dual-pathway architecture that preserves general multimodal capability during embodied specialization~\cite{TwinBrainVLA_2026_arXiv}, and a language-grounded training objective that reduces the tendency of VLA policies to rely only on visual context~\cite{LangForce_2026_arXiv}.

\begin{figure*}[t]
    \centering
    \includegraphics[width=\textwidth]{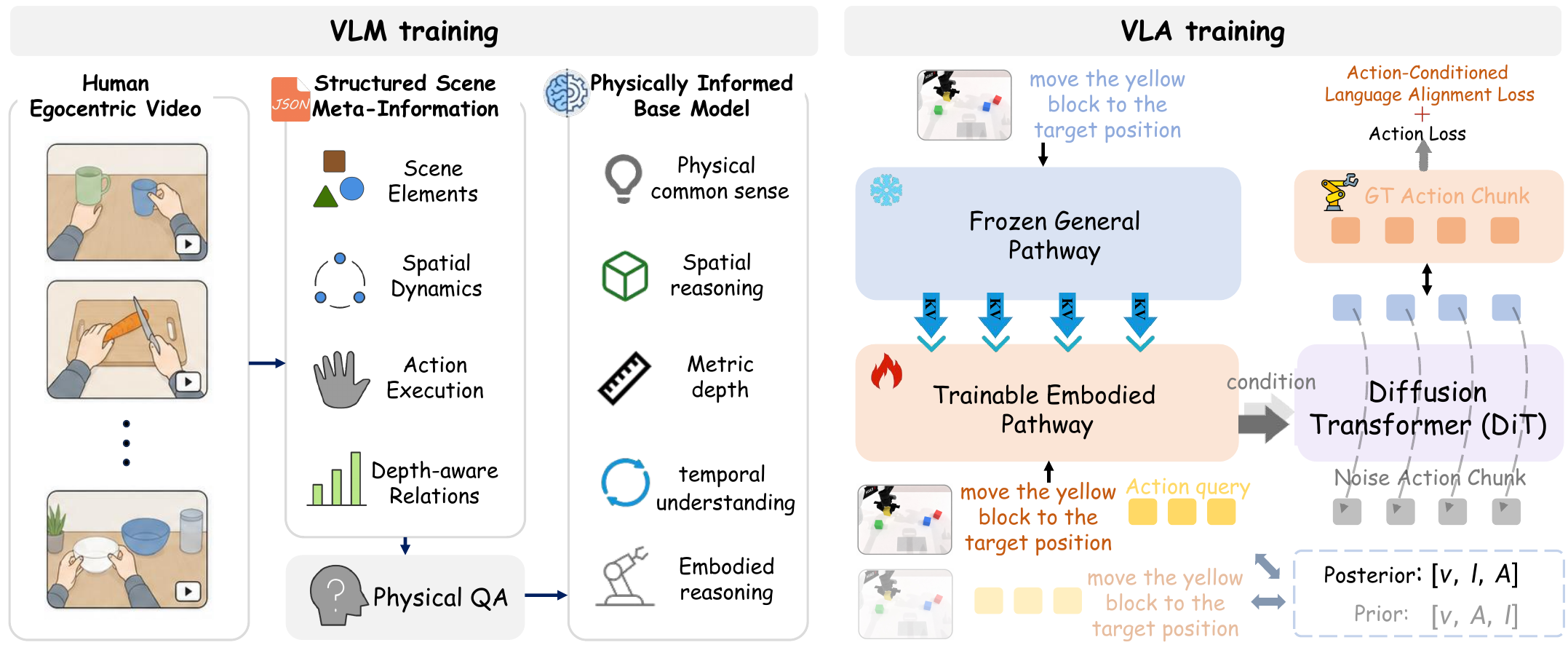}
    \caption{\textbf{Overview of the PhysBrain 1.0 training pipeline.}
    PhysBrain 1.0 first converts human egocentric videos into structured scene meta-information, including scene elements, spatial dynamics, action execution, and depth-aware relations. These records are rendered into physically grounded QA supervision to train a physically informed base model with stronger physical commonsense, spatial reasoning, metric depth understanding, temporal understanding, and embodied reasoning. The resulting base model is then adapted to VLA control through a dual-pathway architecture: a frozen general pathway preserves broad multimodal capability, while a trainable embodied pathway learns continuous action generation. During VLA training, posterior action queries condition on both vision and language, whereas prior action queries are constructed without direct language access; the resulting action-conditioned language alignment loss is optimized together with the action loss for the diffusion-transformer action decoder.}
    \label{fig:physbrain_overview}
\end{figure*}

The architecture is therefore organized around a practical constraint: robot adaptation should specialize the model for control without discarding the multimodal and physical priors learned earlier. PhysBrain 1.0 consists of three coupled components. First, a physically informed base VLM is trained with the QA supervision generated by the data engine. Second, a dual-pathway VLA adaptation module keeps a stable general pathway while training a separate embodied pathway for action. Third, a language-aware action objective and a flow-matching action decoder adapt the model to continuous robot control while keeping the policy sensitive to instructions. Figure~\ref{fig:physbrain_overview} summarizes this full pipeline, from human-video-derived physical QA supervision to capability-preserving VLA adaptation.

\subsection{Physically Informed Base Model}

PhysBrain 1.0 starts from a general multimodal backbone and adapts it with the supervision generated by the data engine. This stage is not a robot-control stage. Its purpose is to improve the base model's ability to interpret first-person physical scenes, including object state, spatial layout, depth, feasible interaction, temporal dynamics, and multi-step task structure.

The base model is trained on free-form QA examples whose content is grounded in structured scene meta-information. For first-person physical reasoning examples, the answer format follows a perception-state-planning-execution organization: the model first identifies the environment and task-relevant objects, then reasons about physical state and spatial relations, and finally describes a feasible plan or execution process. For retention and generality, this training is mixed with broader multimodal QA families such as OCR, chart reasoning, visual logic, and domain knowledge.

This stage defines what is transferred into VLA adaptation. Rather than asking robot demonstrations to teach all physical regularities from scratch, PhysBrain 1.0 first trains a VLM to organize scenes in terms of objects, relations, state changes, depth, and task feasibility. Robot data are then used to map these priors onto a particular embodiment and action space.

\subsection{Capability Preservation During Embodied Adaptation}

Adapting a VLM into a robot controller introduces a tension between two objectives. The model must learn low-level action prediction from robot demonstrations, but the optimization should not overwrite the general multimodal representations that make the model useful for open-ended perception and instruction understanding. PhysBrain 1.0 addresses this tension with two coordinated pathways during VLA training.

The general pathway is initialized from the physically informed base VLM and kept frozen during robot adaptation. It processes the visual observation and language instruction as a stable semantic reference. The embodied pathway is initialized from the same model family but remains trainable; it receives the task context used for action prediction and is optimized on robot demonstrations. This separation gives the control pathway access to general semantic features while avoiding direct updates to the frozen reference pathway.

The two pathways communicate through asymmetric layer-wise fusion. Let $\mathbf{H}_G^l$ and $\mathbf{H}_E^l$ denote the hidden states of the general and embodied pathways at layer $l$. The general pathway updates only through its frozen self-attention and feed-forward blocks. The embodied pathway computes its query from $\mathbf{H}_E^l$, while its key-value context concatenates its own states with stop-gradient features from the general pathway:
\begin{align}
    K_{\mathrm{joint}}^l &= [\mathrm{sg}(K_G^l); K_E^l], \\
    V_{\mathrm{joint}}^l &= [\mathrm{sg}(V_G^l); V_E^l], \\
    \mathbf{H}_E^{l+1} &= \mathrm{Attn}(Q_E^l, K_{\mathrm{joint}}^l, V_{\mathrm{joint}}^l) + \mathrm{FFN}_E(\mathbf{H}_E^l),
\end{align}
where $\mathrm{sg}(\cdot)$ denotes stop-gradient. This asymmetric interaction lets the embodied pathway condition on preserved semantic information, while gradients from action learning update only the trainable control pathway and action decoder. The design reduces the pressure on a single parameter set to both retain broad multimodal competence and specialize for motor control.

\subsection{Action-Conditioned Language Alignment}

Capability preservation alone does not guarantee instruction following. This issue becomes especially relevant under a data-efficient robot adaptation setting. Because PhysBrain 1.0 uses limited robot trajectories for downstream adaptation, the robot-side data can be much narrower than the large-scale multimodal data used to build the base model. In goal-conditioned robot datasets, such narrowness can make the instruction highly predictable from the scene: a particular object arrangement may appear with only a small set of language commands. In this setting, a policy trained only with action imitation can learn a visually driven shortcut and use the instruction weakly, especially under out-of-distribution language or task composition.

PhysBrain 1.0 uses action queries to compare a vision-only action context with a language-conditioned action context. Let $v$ denote visual tokens, $\ell$ denote the language instruction, and $\mathcal{A}$ denote learnable action query tokens. The prior branch arranges the sequence as
\begin{equation}
    \mathrm{Input}_{\mathrm{prior}} = [v, \mathcal{A}, \ell],
\end{equation}
so the causal action queries can attend to vision but not to the instruction. The posterior branch arranges the sequence as
\begin{equation}
    \mathrm{Input}_{\mathrm{post}} = [v, \ell, \mathcal{A}],
\end{equation}
so the action queries can attend to both vision and language. The hidden states of these query tokens provide the conditions for action prediction in the two branches.

The paired branches support a log-likelihood-ratio style objective. The prior branch estimates how much language can be explained from vision and action-query information, while the posterior branch provides the language-conditioned action representation. This comparison objective encourages the action representation to retain information that is relevant to the instruction, instead of relying only on correlations between visual observations and actions. In practice, this term is optimized together with the action prediction losses, with stop-gradient operations used to prevent the baseline language model from being degraded to increase the ratio.

This objective is used for a specific purpose: to make robot adaptation depend on instructions when the instruction changes the intended action. It complements the dual-pathway architecture. The frozen general pathway preserves semantic competence, while the language-aware action objective encourages the trainable embodied pathway to use that semantics during control learning.

\subsection{Unified Action Generation}

PhysBrain 1.0 decodes continuous robot actions from the hidden states of the language-conditioned action queries. The action decoder is trained with a flow-matching objective. Let $\mathbf{a}_1$ be the ground-truth action trajectory, $\mathbf{a}_0 \sim \mathcal{N}(0, I)$ be Gaussian noise, and $\mathbf{a}_t = (1-t)\mathbf{a}_0 + t\mathbf{a}_1$ be the interpolated action at time $t$. Given a query-state condition $\mathbf{C}$ from the embodied pathway, the decoder predicts the velocity field:
\begin{equation}
    \mathcal{L}_{\mathrm{FM}}(\psi; \mathbf{C}) =
    \mathbb{E}_{t,\mathbf{a}_0,\mathbf{a}_1}
    \left[
    \left\| v_\psi(\mathbf{a}_t, t, \mathbf{C}) - (\mathbf{a}_1 - \mathbf{a}_0) \right\|_2^2
    \right].
\end{equation}

The predicted trajectory is represented in an end-effector-frame (EEF) action space, including translational and rotational components. This action representation is consistent with the motivation for metric depth QA in the data engine: understanding absolute distance and displacement in the visual world provides useful structure for predicting continuous pose changes. During inference, PhysBrain 1.0 uses the posterior branch, conditions the action decoder on the language-aware action query states, and generates continuous control commands.

\subsection{Robot Adaptation Protocol and Data Efficiency}

The final stage adapts PhysBrain 1.0 to concrete robot benchmarks using benchmark-specific robot trajectories. SimplerEnv-WidowX uses Bridge data~\cite{Bridgedatav2_2023_CoRL}; SimplerEnv-GoogleRobot uses factual data~\cite{SimplerEnv_2024_CoRL}; LIBERO uses the standard spatial, object, goal, and long-horizon settings~\cite{OpenVLA_2024_CoRL}; and RoboCasa-GR1 uses PhysicalAI-Robotics-GR00T-X-Embodiment-Sim for embodiment adaptation~\cite{RoboCasa_2024_RSS,GR00T_2025_arXiv}.

PhysBrain 1.0 does not remove the need for robot data. It changes the role of robot data. Human first-person video supplies a large portion of the physical and spatial prior, while robot trajectories teach the model how those priors map onto a specific embodiment, action parameterization, and benchmark distribution. The expected benefit is data efficiency: if the model already understands object state, reachability, spatial layout, metric distance, and instruction-conditioned task structure, fewer robot demonstrations are needed to adapt those priors to action.

The architecture therefore completes the training logic introduced earlier. The data engine extracts physical knowledge from human video; the base VLM internalizes this knowledge through physically grounded QA; the dual-pathway adaptation module preserves general multimodal capability while learning robot control; the language-aware action objective helps maintain instruction sensitivity during imitation learning; and the final robot adaptation stage uses a small amount of target-domain robot data to map these priors efficiently into concrete control policies.

\section{Experiments}
\label{sec:experiments}



\subsection{VLM Experiments}
\label{subsec:vlm_exp}

We first evaluate whether the physical supervision produced by the PhysBrain 1.0 data engine improves the multimodal reasoning ability of the base model before robot adaptation. Starting from Qwen3-VL~\cite{Qwen3-VL_2025_arXiv}, we train PhysBrain 4B and PhysBrain 8B with the large-scale QA data generated from our scene meta-information pipeline. The goal of this stage is not to optimize for a single benchmark, but to strengthen physically grounded visual understanding while preserving broad multimodal competence.

\subsubsection{VLM Experiment Settings}

We evaluate PhysBrain 4B and PhysBrain 8B on seven visual question-answering benchmarks: ERQA~\citep{GoogleRobotics_2025_arxiv}, PhysBench~\citep{PhysBench_2025_arXiv}, MME~\cite{MME_2023_arXiv}, MMMU~\citep{MMMU_2024_CVPR}, OCRBench~\citep{OCRBenchV2_2025_arXiv}, RealWorldQA~\citep{OCRBenchV2_2025_arXiv}, and TextVQA~\citep{TextVQA_2019_CVPR}. ERQA and PhysBench focus more directly on embodied and physical reasoning, while MME, MMMU, OCRBench, RealWorldQA, and TextVQA measure general multimodal perception, knowledge reasoning, OCR, and real-world visual understanding. We compare against the corresponding Qwen3-VL base models and representative recent multimodal baselines, using the official evaluation protocols for each benchmark.

\subsubsection{VLM Experiment Results}

\begin{figure*}[t]
    \centering
    \includegraphics[width=\textwidth]{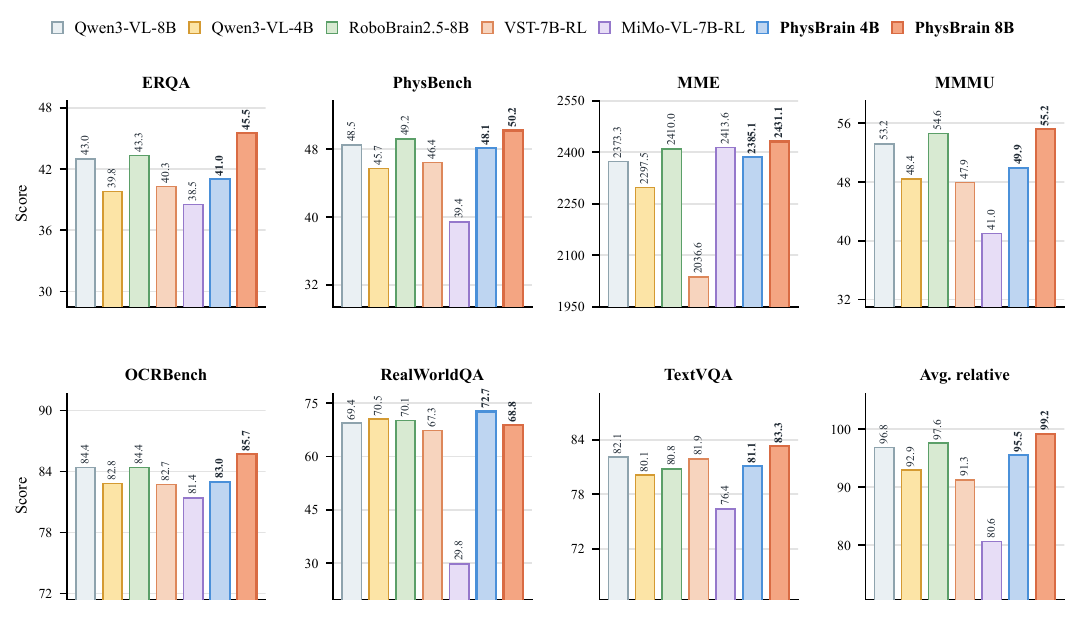}
    \caption{\textbf{Multimodal question-answering benchmark results.}
    The first seven panels report the raw score on each benchmark. The final ``Avg. relative'' panel is not an average of raw scores; for each benchmark, we first divide every model's score by the best score on that benchmark and convert it to a percentage, then average these seven relative percentages. Each panel uses an independent y-axis range to make within-benchmark differences visible, with the exact score annotated above each bar. PhysBrain models are highlighted with the blue and peach accents used in the overview figure, and higher values are better for all benchmarks.}
    \label{fig:vlm_qa_results}
\end{figure*}

As shown in Figure~\ref{fig:vlm_qa_results}, PhysBrain 8B achieves the strongest overall performance profile among the compared models, obtaining the best scores on ERQA, PhysBench, MME, MMMU, OCRBench, and TextVQA, while PhysBrain 4B obtains the best score on RealWorldQA. Compared with Qwen3-VL-8B, PhysBrain 8B improves from 43.0 to 45.5 on ERQA, from 48.5 to 50.2 on PhysBench, from 2373.3 to 2431.1 on MME, and from 53.2 to 55.2 on MMMU. These gains indicate that the PhysBrain data engine improves both physically grounded reasoning and general multimodal capability rather than trading one for the other.

PhysBrain 4B also consistently improves over Qwen3-VL-4B across all reported benchmarks, including a large gain on RealWorldQA from 70.5 to 72.7. This suggests that the benefit of our physically grounded QA supervision is not limited to the larger model scale. Overall, the VLM results support the central design of PhysBrain 1.0: before adapting to robot control, the model first acquires stronger visual, spatial, and physical commonsense from human egocentric data, providing a more capable foundation for downstream VLA training.

\subsection{VLA Simulation Experiments}
\label{subsec:vla_exp}

To evaluate the embodied control ability of PhysBrain 1.0, we conduct VLA simulation experiments on four benchmark settings from three benchmark families: SimplerEnv-WidowX, SimplerEnv-GoogleRobot~\cite{SimplerEnv_2024_CoRL}, RoboCasa-GR1~\cite{RoboCasa_2024_RSS, GR00T_2025_arXiv}, and LIBERO~\cite{LIBERO_2023_NeurIPS}. These settings cover different robot embodiments, manipulation tasks, and evaluation protocols. Since embodied benchmarks are tied to different robot morphologies, controllers, observation spaces, and action conventions, each benchmark setting requires VLA training on robot data from the corresponding embodiment. This setting tests whether PhysBrain 1.0 can be adapted across multiple embodied data regimes rather than only within a single robot platform.

\subsubsection{VLA Experiment Settings}

\paragraph{Training}
To adapt PhysBrain 1.0 from a VLM into a VLA policy, we use a unified post-training recipe across our simulation experiments. For each benchmark setting, the model is fine-tuned with the corresponding embodiment-specific robot demonstrations while keeping the PhysBrain architecture and optimization settings consistent across our adaptation runs. This protocol isolates the effect of the pretrained physical priors and benchmark-specific robot adaptation.

\begin{table*}[!t]
  \centering
  \caption{
    \textbf{Results of evaluating the VLA models with the WidowX robot in the SimplerEnv-WidowX simulation benchmark}. We highlight the best results in \textbf{bold} and the second-best results with \underline{underline}.
    }
  \begin{adjustbox}{width=\linewidth}
  \begin{tabular}{l c c c c c}
    \toprule
    \textbf{Method}
     & \makecell[c]{\textbf{Put Spoon} \\ \textbf{on Towel}} 
     & \makecell[c]{\textbf{Put Carrot} \\ \textbf{on Plate}} 
     & \makecell[c]{\textbf{Stack Green Block} \\ \textbf{on Yellow Block}} 
     & \makecell[c]{\textbf{Put Eggplant} \\ \textbf{in Yellow Basket}} 
     & \textbf{Average} \\
    \midrule

    RT-1-X~\citep{OXE_2024_ICRA}         &  0.0  & 4.2   & 0.0   & 0.0   & 1.1 \\
    Octo-Base~\citep{Octo_2024_arXiv}       & 15.8  & 12.5  & 0.0   & 41.7  & 17.5 \\
    Octo-Small~\citep{Octo_2024_arXiv}      & 41.7  & 8.2   & 0.0   & 56.7  & 26.7 \\
    OpenVLA~\citep{OpenVLA_2024_CoRL}         & 4.2   & 0.0   & 0.0   & 12.5   & 4.2 \\
    OpenVLA-OFT~\citep{OpenVLA-OFT_2025_arXiv}     & 12.5  & 4.2  & 4.2  & 72.5  & 23.4 \\
    RoboVLM~\citep{RoboVLM_2024_arXiv}         & 50.0  & 37.5  & 0.0   & 83.3  & 42.7 \\ 
    TraceVLA~\citep{TraceVLA_2025_arXiv}        & 12.5  & 16.6  & 16.6  & 65.0  & 27.7 \\
    SpatialVLA~\citep{Spatialvla_2025_arXiv}      & 20.8  & 20.8  & 25.0  & 70.8  & 34.4 \\
    CogACT~\citep{CogACT_2024_arXiv}          & 71.7 &  50.8  & 15.0 & 67.5 & 51.3 \\
    VideoVLA~\citep{VideoVLA_2025_NeurIPS}        & 75.0 & 20.8   & 45.8 & 70.8 & 53.1 \\
    $\pi_0$~\citep{PI0_2024_arXiv}         & 29.1 & 0.0 & 16.6 & 62.5 & 27.1 \\
    $\pi_{0.5}$~\citep{PI05_2025_arXiv} & 49.3 & 64.7 & 44.7 & 69.7 & 57.1 \\
    Isaac-GR00T-N1.6-Bridge~\citep{GR00T_N1.6}   & 64.5 & 65.5 & 5.5 & 93.0 & 57.1 \\
    Xiaomi-Robotics-0~\citep{XiaomiRobotics0_2026_arxiv} &  95.8 &  62.5 &  75.0 &  83.3 &  \underline{79.2} \\
    
    \cmidrule(lr){1-6}

    \rowcolor{gray!30}\textbf{PhysBrain 1.0} (ours) & 95.8 & 65.5 & 59.4 &  100.0  & \textbf{80.2} \\
    \bottomrule
  \end{tabular}
  \end{adjustbox}
  \vspace{0.5 em}
  \label{tab:simplerenv_widowx_results}
\end{table*}

\begin{table*}[!t]
  \centering
  \caption{
    \textbf{Results of evaluating the VLA models with the Google Robot in the SimplerEnv-GoogleRobot simulation benchmark}. We highlight the best results in \textbf{bold} and the second-best results with \underline{underline}.
    }
    \setlength{\tabcolsep}{15pt}
  \begin{adjustbox}{width=0.82\linewidth}
  \begin{tabular}{l c c c c}
    \toprule
    \textbf{Method}
     & \makecell[c]{\textbf{Pick} \\ \textbf{Coke Can}}
     & \makecell[c]{\textbf{Move} \\ \textbf{Near}}
     & \makecell[c]{\textbf{Open/Close} \\ \textbf{Drawer}}
     & \textbf{Average} \\
    \midrule
    $\pi_0$~\citep{PI0_2024_arXiv} & 75.2 & 63.7 & 25.6 & 54.8 \\
    GR00T-N1~\citep{GR00T_2025_arXiv} & 78.8 & 62.5 & 13.2 & 51.5 \\
    GreenVLA (R1)~\citep{GreenVLA_2026_arxiv} & 90.4 & 61.2 & 62.9 & 66.9 \\
    X-VLA~\citep{X-VLA_2025_arXiv} & 85.5 & 79.8 & 61.9 & 75.7 \\
    Xiaomi-Robotics-0~\citep{XiaomiRobotics0_2026_arxiv} & 98.7 & 88.8 & 79.6 & \underline{89.03} \\
    \midrule
    \rowcolor{gray!30}\textbf{PhysBrain 1.0} (ours) & 100.0 & 94.8 & 79.2 & \textbf{91.33} \\
    \bottomrule
  \end{tabular}
  \end{adjustbox}
  \vspace{0.5 em}
  \label{tab:simplerenv_googlerobot_results}
\end{table*}

\paragraph{Benchmarks}
SimplerEnv evaluates manipulation policies in simulation across multiple robot embodiments~\cite{SimplerEnv_2024_CoRL}. We report results on both SimplerEnv-WidowX and SimplerEnv-GoogleRobot. The two settings are trained and evaluated separately with embodiment-specific training data, rather than being used as a cross-embodiment transfer test. For SimplerEnv-WidowX, we train on the BridgeV2 real-robot dataset and evaluate on four held-out simulation tasks, making the benchmark a test of out-of-domain generalization. The results are shown in Table~\ref{tab:simplerenv_widowx_results}. For SimplerEnv-GoogleRobot, we train with Google Robot adaptation data and evaluate on the out-of-domain Pick Coke Can, Move Near, and Open/Close Drawer tasks with the Google Robot embodiment. The results are shown in Table~\ref{tab:simplerenv_googlerobot_results}.

RoboCasa-GR1 is a tabletop manipulation benchmark built on RoboCasa~\cite{RoboCasa_2024_RSS}, where a GR1 robot performs bimanual manipulation with two dexterous hands. We evaluate on 24 tabletop tasks and train with the 24K GR1 teleoperation simulation demonstrations released by NVIDIA. This benchmark tests multi-task VLA learning and dexterous-hand control. The results are shown in Table~\ref{tab:robocasa_results}.

LIBERO is a Franka-based simulation benchmark for language-conditioned manipulation~\cite{LIBERO_2023_NeurIPS}. We evaluate on four task suites and train with the official expert demonstrations provided by the benchmark. LIBERO complements RoboCasa-GR1 and the two SimplerEnv embodiments by testing PhysBrain 1.0 on a standardized single-arm embodiment with expert trajectories. The results are shown in Table~\ref{tab:libero_results}.

\subsubsection{VLA Experiment Results}

\begin{table*}[!t]
    \centering
    \small
    \renewcommand{\arraystretch}{1.3}
    \setlength{\tabcolsep}{3pt}

    \caption{
      \textbf{Results of evaluating the VLA models with the GR1 robot in the RoboCasa Tabletop simulation environment}. The results for QwenGR00T, QwenOFT, and QwenFAST are derived from the official StarVLA experiments~\citep{starvla_2025}. We highlight the best results in \textbf{bold} and the second-best results with \underline{underline}.
    }
    \begin{adjustbox}{width=\textwidth}
    \begin{tabular}{l c c c c c c}
        \toprule
        \textbf{Task} & 
        \textbf{\scriptsize \makecell{Isaac-GR00T\\N1.6}} & 
        \textbf{\scriptsize \makecell{QwenGR00T\\ + Qwen3VL}} &  
        \textbf{\scriptsize \makecell{QwenOFT\\ + Qwen3VL}} & 
        \textbf{\scriptsize \makecell{QwenFAST\\ + Qwen3VL}} & 
        \textbf{\scriptsize \makecell{VP-VLA}} & 
        \textbf{\scriptsize \makecell{PhysBrain 1.0}} \\
        \midrule
        \rowcolors{1}{gray!15}{white}
        PnP Bottle To Cabinet Close & 51.5 & 46.0 & 30.0 & 38.0 & 54.0 & 76.0 \\
        PnP Can To Drawer Close & 13.0 & 80.0 & 76.0 & 44.0 & 72.0 & 78.0 \\
        PnP Cup To Drawer Close & 8.5 & 54.0 & 44.0 & 56.0 & 44.0 & 66.0 \\
        PnP Milk To Microwave Close & 14.0 & 48.0 & 44.0 & 44.0 & 74.0 & 60.0 \\
        PnP Potato To Microwave Close & 41.5 & 28.0 & 32.0 & 14.0 & 34.0 & 60.0 \\
        PnP Wine To Cabinet Close & 16.5 & 46.0 & 36.0 & 14.0 & 48.0 & 56.0 \\
        PnP Novel From Cuttingboard To Basket & 58.0 & 48.0 & 50.0 & 54.0 & 66.0 & 58.0 \\
        PnP Novel From Cuttingboard To Cardboardbox & 46.5 & 40.0 & 40.0 & 42.0 & 54.0 & 60.0 \\
        PnP Novel From Cuttingboard To Pan & 68.5 & 68.0 & 70.0 & 58.0 & 74.0 & 80.0 \\
        PnP Novel From Cuttingboard To Pot & 65.0 & 52.0 & 54.0 & 58.0 & 54.0 & 66.0 \\
        PnP Novel From Cuttingboard To Tieredbasket & 46.5 & 56.0 & 38.0 & 40.0 & 56.0 & 62.0 \\
        PnP Novel From Placemat To Basket & 58.5 & 42.0 & 32.0 & 36.0 & 48.0 & 54.0 \\
        PnP Novel From Placemat To Bowl & 57.5 & 44.0 & 58.0 & 38.0 & 74.0 & 72.0 \\
        PnP Novel From Placemat To Plate & 63.0 & 48.0 & 52.0 & 42.0 & 70.0 & 74.0 \\
        PnP Novel From Placemat To Tieredshelf & 28.5 & 18.0 & 24.0 & 18.0 & 26.0 & 18.0 \\
        PnP Novel From Plate To Bowl & 57.0 & 60.0 & 60.0 & 52.0 & 52.0 & 76.0 \\
        PnP Novel From Plate To Cardboardbox & 43.5 & 50.0 & 50.0 & 30.0 & 44.0 & 68.0 \\
        PnP Novel From Plate To Pan & 51.0 & 54.0 & 66.0 & 48.0 & 56.0 & 76.0 \\
        PnP Novel From Plate To Plate & 78.7 & 70.0 & 68.0 & 50.0 & 62.0 & 78.0 \\
        PnP Novel From Tray To Cardboardbox & 51.5 & 38.0 & 44.0 & 28.0 & 44.0 & 72.0 \\
        PnP Novel From Tray To Plate & 71.0 & 56.0 & 56.0 & 34.0 & 66.0 & 80.0 \\
        PnP Novel From Tray To Pot & 64.5 & 50.0 & 62.0 & 46.0 & 38.0 & 70.0 \\
        PnP Novel From Tray To Tieredbasket & 57.0 & 36.0 & 54.0 & 36.0 & 58.0 & 52.0 \\
        PnP Novel From Tray To Tieredshelf & 31.5 & 16.0 & 30.0 & 16.0 & 24.0 & 36.0 \\
        \hiderowcolors
        \midrule
        \rowcolor{gray!30}
        \textbf{Average} & 47.6 & 47.8 & 48.8 & 39.0 & \underline{53.8} & \textbf{64.5} \\
        \bottomrule
    \end{tabular}
    \end{adjustbox}
    \label{tab:robocasa_results}
\end{table*}

\begin{table}[!t]
    \centering
    \setlength{\tabcolsep}{15pt}
    \renewcommand{\arraystretch}{1}
    \caption{
    LIBERO simulation results on four task suites.
    We report success rates (\%) on Spatial, Object, Goal, and Long, together with the average across the four suites.
    The first block lists representative policy/VLA systems, while the final block isolates the controlled comparison between full-frame training and PhysBrain 1.0.
    }
    \begin{adjustbox}{max width=0.92\linewidth}
    \begin{tabular}{lccccc}
        \toprule
        \textbf{Method} & \textbf{L-Spatial} & \textbf{L-Object} & \textbf{L-Goal} & \textbf{L-Long} & \textbf{Avg.} \\
        \midrule
        Diffusion Policy~\citep{DiffusionPolicy_23} & 78.5 & 87.5 & 73.5 & 64.8 & 76.1 \\
        OpenVLA~\citep{OpenVLA_2024_CoRL}          & 84.7 & 88.4 & 79.2 & 53.7 & 76.5 \\
        SpatialVLA~\citep{Spatialvla_2025_arXiv}       & 88.2 & 89.9 & 78.6 & 55.5 & 78.1 \\
        CoT-VLA~\citep{CoT-VLA}          & 87.5 & 91.6 & 87.6 & 69.0 & 83.9 \\
        GR00T N1~\citep{GR00T_2025_arXiv}         & 94.4 & 97.6 & 93.0 & 90.6 & 93.9 \\
        F1~\citep{F1-VLA_2025_arXiv}               & 98.2 & 97.8 & 95.4 & 91.3 & 95.7 \\
        InternVLA-M1~\citep{InternVLA_M1_25}     & 98.0 & 99.0 & 93.8 & 92.6 & 95.9 \\
        $\pi_0$~\citep{PI0_2024_arXiv}          & 98.0 & 96.8 & 94.4 & 88.4 & 94.4 \\
        $\pi_{0.5}$~\citep{PI05_2025_arXiv}      & 98.8 & 98.2 & 98.0 & 92.4 & 96.9 \\
        GR00T N1.6~\citep{GR00T_N1.6}       & 97.7 & 98.5 & 97.5 & 94.4 & 97.0 \\
        Xiaomi-Robotics-0~\citep{XiaomiRobotics0_2026_arxiv} & 98.8 & 100.0 & 98.8 & 97.2 & \underline{98.7} \\
        \midrule
        \rowcolor{gray!30}\textbf{PhysBrain 1.0 (ours)} & 99.6 & 99.6 & 99.4 & 96.4 & \textbf{98.8} \\
        \bottomrule
    \end{tabular}
    \end{adjustbox}
    \label{tab:libero_results}
\end{table}

\paragraph{SimplerEnv-WidowX}
As shown in Table~\ref{tab:simplerenv_widowx_results}, PhysBrain 1.0 obtains the best average success rate on the SimplerEnv-WidowX benchmark, reaching 80.2\% across the four held-out tasks. This is 1.0 percentage point above the strongest prior method, Xiaomi-Robotics-0, and 23.1 percentage points above both $\pi_{0.5}$ and Isaac-GR00T-N1.6-Bridge. The task-level results show that PhysBrain 1.0 is not driven by a single easy category: it ties the best result on \emph{Put Spoon on Towel}, ties the best result on \emph{Put Carrot on Plate}, reaches 100.0\% on \emph{Put Eggplant in Yellow Basket}, and remains competitive on the block-stacking task. Since this setting trains on BridgeV2 data but evaluates on SimplerEnv simulation tasks, the result suggests that the PhysBrain prior improves out-of-domain generalization for the WidowX embodiment.

\paragraph{SimplerEnv-GoogleRobot}
As shown in Table~\ref{tab:simplerenv_googlerobot_results}, PhysBrain 1.0 also achieves the best average result on SimplerEnv-GoogleRobot, improving the average success rate to 91.33\%. Compared with the strongest baseline, Xiaomi-Robotics-0, PhysBrain 1.0 improves by 2.30 percentage points on average. The gain is clearest on \emph{Move Near}, where PhysBrain 1.0 improves from 88.8\% to 94.8\%, while also reaching 100.0\% on \emph{Pick Coke Can}. On \emph{Open/Close Drawer}, PhysBrain 1.0 remains comparable to the strongest baseline. Together with the WidowX result, this shows that the same PhysBrain training recipe transfers to two distinct SimplerEnv embodiments rather than only fitting a single robot platform.

\paragraph{RoboCasa-GR1}
As shown in Table~\ref{tab:robocasa_results}, PhysBrain 1.0 achieves the strongest average performance on RoboCasa-GR1, reaching 64.5\% across 24 tabletop manipulation tasks. This is 10.7 percentage points above VP-VLA, the second-best method in the table, and 15.7 percentage points above QwenOFT with Qwen3VL. The improvement is important because RoboCasa-GR1 differs from the SimplerEnv settings in both embodiment and task structure: the benchmark uses a GR1 robot with bimanual dexterous hands and a broad set of pick-and-place tasks. The result indicates that the PhysBrain pretraining signal remains useful after adaptation to dexterous tabletop manipulation, not only to single-arm or mobile manipulation settings.

\paragraph{LIBERO}
As shown in Table~\ref{tab:libero_results}, PhysBrain 1.0 reaches 98.8\% average success on LIBERO, slightly improving over the previous best average result of 98.7\% from Xiaomi-Robotics-0. LIBERO is already close to saturation for several recent VLA systems, so the margin is smaller than in SimplerEnv and RoboCasa-GR1. Nevertheless, PhysBrain 1.0 achieves the best average score while remaining strong across all four suites, including 99.6\% on L-Spatial and 99.4\% on L-Goal. This result shows that the method does not trade off standardized single-arm imitation performance for gains on the more out-of-domain or dexterous benchmarks.

\paragraph{Summary}
Across the four VLA evaluations, PhysBrain 1.0 achieves the best average score in every reported table. The largest gains appear on RoboCasa-GR1 and the two SimplerEnv settings, where the benchmark distribution differs substantially from the training data or embodiment-specific adaptation is more challenging. On LIBERO, where recent systems already approach saturation, PhysBrain 1.0 still matches or slightly exceeds the strongest prior results. These results support the central experimental claim: physical priors learned from structured human-video supervision improve downstream VLA adaptation across heterogeneous embodiments, task distributions, and evaluation protocols.

\section{Real-World Experiments}
\label{sec:real_world_franka}

To validate the transferability of PhysBrain 1.0's physical priors to real-world robot control, we conducted extensive experiments on a Franka Research 3 robot arm equipped with a Robotiq 2F-85 parallel-jaw gripper. The experiments focus on tabletop vegetable grasping tasks, which require fine-grained physical understanding of object geometry, material properties, and contact dynamics.

\paragraph{Experimental Setup}
The robot is mounted in front of a table where various vegetables are placed. The workspace includes common items such as eggplants, carrots, cucumbers, potatoes, tomatoes, romaine lettuce, and Chinese cabbage. Each object category presents distinct physical challenges: smooth surfaces (eggplants, tomatoes), irregular shapes (carrots, potatoes), deformable structures (romaine lettuce, Chinese cabbage), and varying stiffness (cucumbers vs.~ripe tomatoes). Two Intel RealSense D435i cameras provide RGB observations: one mounted as an external viewpoint overlooking the workspace, and another mounted on the robot wrist for close-up observation during manipulation.

\begin{figure}[htbp]
    \centering
    \begin{minipage}[c]{0.35\linewidth}
        \centering
        \includegraphics[width=\linewidth]{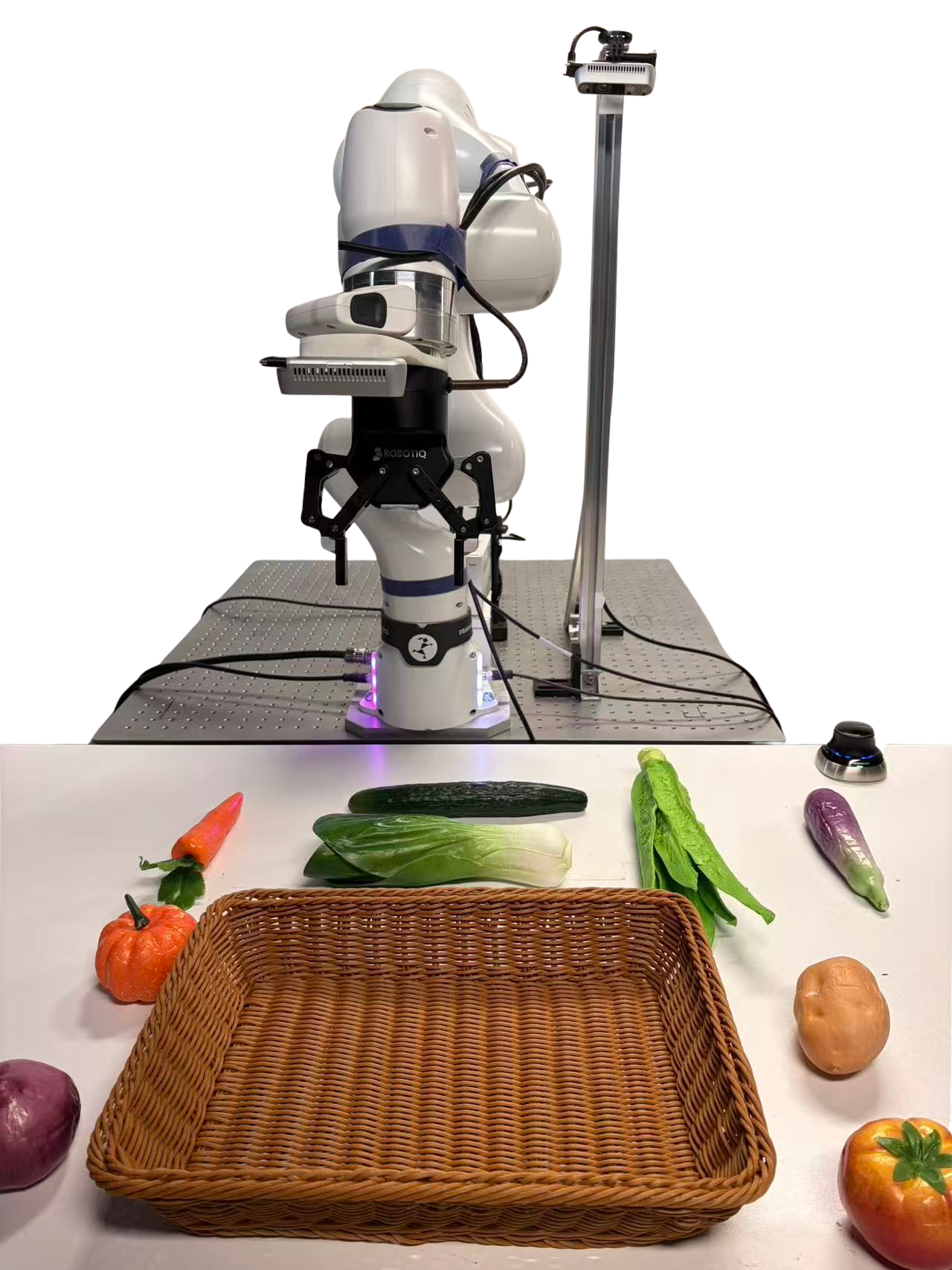}
        \\[-1mm]
        \small (a) Front view
    \end{minipage}
    \hfill
    \begin{minipage}[c]{0.62\linewidth}
        \centering
        \includegraphics[width=\linewidth]{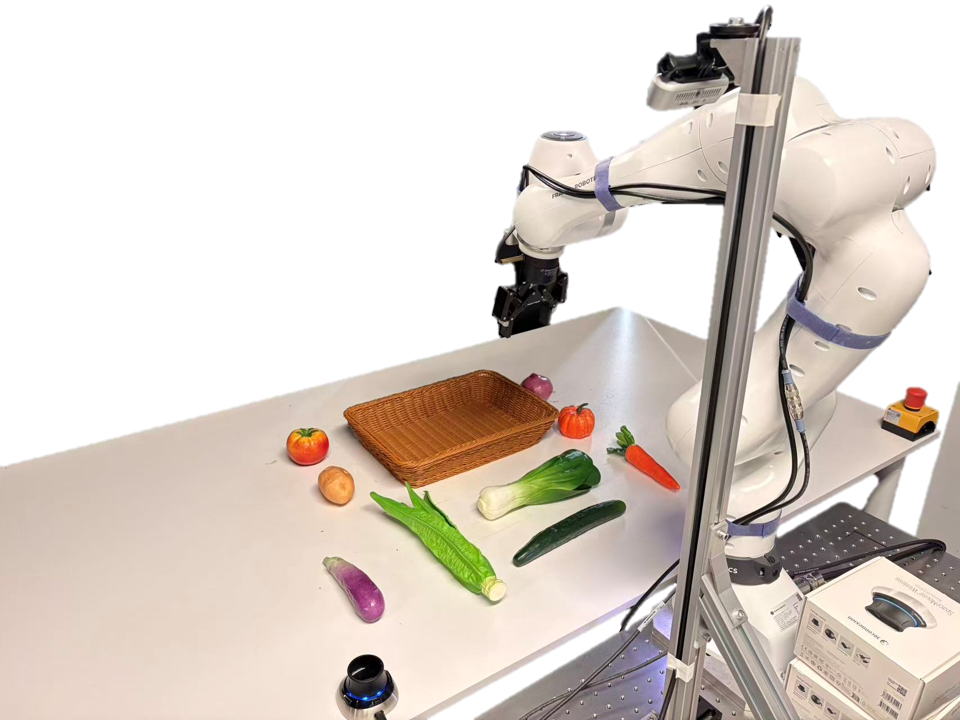}
        \\[-1mm]
        \small (b) Rear-side view
    \end{minipage}
    \caption{\textbf{Real-world experimental setup overview.} The Franka Research 3 robot with a Robotiq 2F-85 gripper is positioned in front of a tabletop workspace with various vegetables. Two Intel RealSense D435i cameras provide RGB observations: one external viewpoint and one wrist-mounted viewpoint for close-up observation during manipulation. (a) Front view of the robot arm. (b) Rear-side view showing the workspace layout.}
    \label{fig:franka_scene_overview}
\end{figure}

\paragraph{Data Collection}
For each vegetable category, we collected 50 demonstration trajectories using a SpaceMouse to control the 6-DoF end-effector pose. The human operator guides the robot arm to grasp the object from various initial poses and orientations, ensuring coverage of diverse grasp configurations. All demonstrations are recorded in the LeRobot 3.0 data format, which provides a standardized structure for robot learning datasets. Each trajectory includes end-effector poses, gripper states, and synchronized RGB observations from both external and wrist-mounted cameras. In total, we collected 450 trajectories across 9 object categories: Chinese cabbage, carrot, cucumber, eggplant, onion, potato, pumpkin, tomato, and romaine lettuce.

\paragraph{Post-Training Protocol}
We performed post-training on the collected real-world data to adapt PhysBrain 1.0 to the Franka embodiment. The post-training follows the same dual-pathway architecture described in Section~\ref{sec:architecture}, with the general pathway frozen and the embodied pathway fine-tuned on the Franka trajectory distribution. The flow-matching action decoder is optimized to predict continuous end-effector motions in the Franka action space. All real-world evaluations use a single post-trained policy across object categories and long-horizon instructions, rather than training separate specialist models for individual vegetables or tasks.

\paragraph{Evaluation Metrics}
We report task success rate over 50 independent trials per task. For single-object grasping, a trial is counted as successful when the robot grasps and lifts the target object into a stable hold. For long-horizon tasks, a trial is counted as successful only when the policy completes the full instruction over the requested set of vegetables.

\paragraph{Long-Horizon Tasks}
In addition to single-object grasping, we evaluate the model on long-horizon abstract tasks that require multi-step reasoning and instruction following. For example, given the instruction ``pick up all the green vegetables and put them into the brown basket,'' the model must identify green vegetables (Chinese cabbage, cucumber, and romaine lettuce), plan a sequence of grasping and placing actions, and execute them in the correct order. We also evaluate an orange-vegetable instruction involving pumpkin and carrot. These tasks test the model's ability to decompose high-level instructions into executable action sequences while maintaining spatial awareness of the scene.

\subsection{Baseline Comparisons}
We compare PhysBrain 1.0 against $\pi_{0.5}$~\cite{PI05_2025_arXiv}, a vision-language-action flow model pre-trained on large-scale robot demonstrations. Both models are post-trained on the same Franka demonstration data and evaluated under the same 50-trial protocol. This controlled comparison isolates whether the physical priors learned by PhysBrain 1.0 before robot adaptation improve real-world manipulation after both systems see the same embodiment-specific data.

\begin{figure*}[t]
    \centering
    \includegraphics[width=\textwidth]{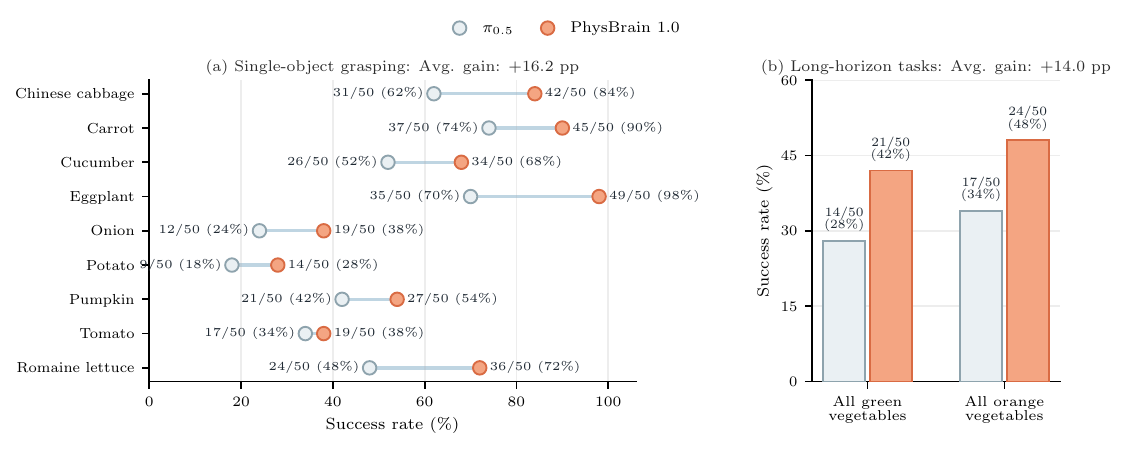}
    \caption{\textbf{Real-world Franka manipulation results.}
    We compare PhysBrain 1.0 with $\pi_{0.5}$ on single-object vegetable grasping and long-horizon semantic instructions. The left panel uses a dumbbell plot to show paired per-category success rates, while the right panel uses vertical bars for the long-horizon tasks. All results are evaluated over 50 trials, with raw success counts annotated next to each mark. All PhysBrain 1.0 results use a single post-trained policy across the evaluated object categories and long-horizon tasks. PhysBrain 1.0 improves the average single-object success rate from 47.1\% to 63.3\% and the average long-horizon success rate from 31.0\% to 45.0\%.}
    \label{fig:real_world_vegetable_results}
\end{figure*}

As shown in Figure~\ref{fig:real_world_vegetable_results}, PhysBrain 1.0 improves over $\pi_{0.5}$ on every evaluated single-object category. Across the nine grasping tasks, $\pi_{0.5}$ succeeds in 212 of 450 trials (47.1\%), while PhysBrain 1.0 succeeds in 285 of 450 trials (63.3\%), corresponding to an average gain of 16.2 percentage points. The gains are especially visible on deformable or visually ambiguous objects such as Chinese cabbage and romaine lettuce, as well as on smooth objects such as eggplant. On the two long-horizon semantic tasks, PhysBrain 1.0 improves from 31 of 100 successful trials (31.0\%) to 45 of 100 successful trials (45.0\%). These results support the central hypothesis of PhysBrain 1.0: human-derived physical priors can improve downstream robot adaptation even when the final policy is trained with the same real-robot demonstrations as a strong VLA baseline.

\section{Discussion}
\label{sec:discussion}

PhysBrain 1.0 is motivated by a change in emphasis rather than by a rejection of imitation learning. Robot trajectories remain necessary for grounding a model in a concrete embodiment, action parameterization, and benchmark distribution. The main difference is that PhysBrain 1.0 does not require these trajectories to carry the entire burden of physical learning. Instead, the system first uses human first-person video to acquire priors about objects, spatial relations, metric distance, state change, action feasibility, and multi-step interaction structure. Robot data are then used primarily to adapt these priors to a target control interface.

This perspective is useful because many physical regularities that matter for control are not robot-specific. When humans see a smooth or slippery object, they naturally anticipate a more secure grasp; when the approach direction is awkward, they adjust wrist orientation before contact; when an object looks fragile, they slow down and reduce impact; and when a handle is partially occluded, they search for a more feasible grasping direction. These are not robot commands, but they are physical priors about contact, friction, reachability, stability, and feasible motion. Human egocentric video contains such priors at scale, and they provide the intermediate organization that can make downstream action learning less sample-intensive. In this sense, physical commonsense acquisition and action imitation play complementary roles: the former shapes what the model understands before control training, while the latter teaches how that understanding should be expressed through robot actions.

The architecture in PhysBrain 1.0follows the same principle. If physical understanding is learned in a general VLM before VLA adaptation, then downstream control training should avoid erasing the capabilities that made the prior useful. The dual-pathway design addresses this by keeping a stable general pathway during robot adaptation, while the trainable embodied pathway learns action prediction. The language-aware action objective addresses a second failure mode: when robot data are limited and scene distributions are narrow, language can become predictable from vision, and a policy may learn to ignore the instruction. Maintaining instruction sensitivity is therefore part of data efficiency, not only an auxiliary alignment objective.


There are also clear limitations. First, the data engine depends on upstream perception and annotation quality. The staged pipeline makes many errors detectable, but it cannot fully eliminate semantic mistakes, missing objects, ambiguous contacts, or incorrect physical interpretations. Second, depth-aware supervision inherits errors from depth estimation and object grounding. The pipeline can detect missing or corrupted depth records, but valid depth maps may still contain local inaccuracies, especially under transparent, reflective, or heavily occluded objects. Third, human egocentric priors are not identical to robot embodiment constraints. Human hands, robot grippers, mobile bases, and simulated manipulators differ in morphology, reachable workspace, force limits, and sensing; robot adaptation is still required to map general physical priors into executable policies.

Finally, benchmark performance should be interpreted within the coverage of the evaluated tasks. SimplerEnv, LIBERO, and RoboCasa test important aspects of manipulation and instruction following, but they do not exhaust long-horizon real-world autonomy, deformable-object interaction, safety-critical execution, or closed-loop recovery under severe distribution shift. Future work should therefore study stronger automatic verification for annotations, better uncertainty handling for depth and grounding, more systematic ablations of human-video supervision, and broader real-robot evaluation. These directions are important for separating the contribution of physical commonsense acquisition from the contribution of benchmark-specific adaptation.

\section{Conclusion}
\label{sec:conclusion}

PhysBrain 1.0 presents a training strategy for embodied foundation models built around the principle of understanding first and action next. Rather than treating larger robot trajectory collections as the only path to stronger embodied control, PhysBrain 1.0 first converts human first-person interaction video into physically grounded supervision and uses it to strengthen the base VLM's understanding of objects, space, depth, dynamics, planning, and execution.

The technical contribution of the data engine is to separate structured scene meta-information from final model supervision. Human video is first parsed into explicit records over scene elements, spatial dynamics, action execution, and depth-aware relations; these records are then used to generate diverse natural-language QA across spatial, temporal, embodied, and general multimodal capabilities. The architecture then transfers these priors into robot control through a physically informed base model, a capability-preserving adaptation design, a language-aware action objective, and a continuous action decoder.

PhysBrain 1.0 emphasizes a pragmatic view of robot data efficiency. Robot trajectories are still essential, but their role shifts from being the sole source of embodied capability to being the adaptation layer that maps human-derived physical priors onto a specific embodiment and action space. This suggests a broader direction for future embodied AI systems: before scaling action imitation, it is important to scale the model's understanding of the physical world in which those actions must be executed.

\section{Contributions}
\label{sec:contributions}

The author's contributions in the following areas are as follows:

\begin{itemize}
    \item \textbf{Data Engine Design:} Xiaopeng Lin, Hang Yuan, Xiaolin Hu, Changti Wu, Yuzhuo Miao, and Yuxuan Tian
    \item \textbf{Data Annotation:} Changti Wu, Yuzhuo Miao, Xiaolin Hu, Hang Yuan, and Shijie Lian
    \item \textbf{Data Quality Control:} Hang Yuan, Xiaolin Hu, Yuzhuo Miao, Xiaopeng Lin and Bin Yu
    \item \textbf{VLA Model Architecture:} Shijie Lian, Bin Yu, and Xiaopeng Lin
    \item \textbf{VLA Training and Evaluation:} Bin Yu, Shijie Lian, Xiaopeng Lin, and Zhaolong Shen
    \item \textbf{VLM Training and Evaluation:} Xiaopeng Lin, Shijie Lian, Bin Yu, and Changti Wu
    \item \textbf{Real-Robot Experiments:} Zhaolong Shen, Xiaopeng Lin, Shijie Lian, and Bin Yu
    \item \textbf{Writing:} Shijie Lian, Bin Yu, Haishan Liu, Zhaolong Shen and Xiaopeng Lin
    \item \textbf{Project Lead:} Kai Chen{\renewcommand{\thefootnote}{\fnsymbol{footnote}}\footnotemark[1]}, Cong Huang and Yukun Shi
\end{itemize}

\begingroup
\renewcommand{\thefootnote}{\fnsymbol{footnote}}
\footnotetext[1]{Corresponding author: \email{kaichen@zgci.ac.cn}.}
\endgroup

{
	\bibliographystyle{plainnat}
	\bibliography{ref}
}

\end{document}